\documentclass[journal]{IEEEtran}
\usepackage{amsfonts}
\usepackage{amssymb}
\usepackage{algorithm2e}
\usepackage{algorithmic}
\usepackage{pgfplots}
\usepackage{mathtools}
\usepackage{algorithm2e}
\usepackage{algorithmic}
\usepackage{pgfplots}
\usepackage{amsfonts}
\usepackage{amssymb}
\usepackage{color}
\usepackage{algorithm2e}
\usepackage{algorithmic}
\usepackage{pgfplots}
\usepackage{amsfonts}
\usepackage{amssymb}
\usepackage{mathtools}
\usepackage{bbm}
\usepackage{dsfont}
\usepackage{float}

\usepackage{amsfonts}
\usepackage{amssymb}
\usepackage{algorithm2e}
\usepackage{algorithmic}
\usepackage{pgfplots}
\usepackage{mathtools}
\usepackage{color}
\usepackage{algorithm2e}
\usepackage{algorithmic}
\usepackage{pgfplots}
\usepackage{amsfonts}
\usepackage{amssymb}
\usepackage{mathtools}
\usepackage{bbm}
\usepackage{dsfont}
\usepackage{float}
\usepackage{hyperref}
\usepackage{comment}

\ifCLASSINFOpdf
\else
\fi
\begin{document}
%
\title{aiTPR: Attribute Interaction-Tensor Product Representation for Image Caption}
%
%
%

\author{Chiranjib~Sur \\ 
		Computer \& Information Science \& Engineering Department, University of Florida.\\
		Email: chiranjib@ufl.edu
}

%
%

\markboth{Journal of XXXX,~Vol.~XX, No.~X, AXX~20XX}%
{Shell \MakeLowercase{\textit{et al.}}: Bare Demo of IEEEtran.cls for IEEE Journals}
%

\maketitle

\begin{abstract}
Region visual features enhance the generative capability of the machines based on features, however they lack proper interaction attentional perceptions and thus ends up with biased or uncorrelated sentences or pieces of misinformation. In this work, we propose Attribute Interaction-Tensor Product Representation (aiTPR) which is a convenient way of gathering more information through orthogonal combination and learning the interactions as physical entities (tensors) and improving the captions. Compared to previous works, where features are added up to undefined feature spaces, TPR helps in maintaining sanity in combinations and orthogonality helps in defining familiar spaces. We have introduced a new concept layer that defines the objects and also their interactions that can play a crucial role in determination of different descriptions. The interaction portions have contributed heavily for better caption quality and has out-performed different previous works on this domain and MSCOCO dataset. We introduced, for the first time, the notion of combining regional image features and abstracted interaction likelihood embedding for image captioning.
\end{abstract}


%
\IEEEpeerreviewmaketitle

\section{Introduction} \label{section:introduction}
\IEEEPARstart{O}{bject} recognition and segmentation has provided ample scope to understand the semantic relationship in images among the different objects \cite{ren2015faster} as a spatio-topological property. This will help in understanding the contexts \cite{Gan2016} and events of the images than mere detection of the objects. The gradual demand and rising interest in industry and AI related frameworks, the requirement is generation and synthesis of reply in structured form. There was a sudden rise in application related to reverse synthesis, caption generation \cite{sur2019representations}, dialogues instead of just prediction of likelihood of decisions. Most of the application of synthesis is based on the requirement to perceive and development of topological dependence of contexts and proceed for generation and synthesis. 
The basis of our image captioning \cite{sur2019survey} model is based on reviving the underlying understanding of the representational aspects from both the context and the role prospective for languages. Here, we define Attribute Interaction-Tensor Product Representation (aiTPR), where the context is more about understanding and the roles play the role of creating the dependency or the interaction \cite{sur2019representations}. Here, we defined a layer that samples the combination, transforms them into a understandable space and help generate the captions. While "English" language is based on structural principles, these are roles and contexts are pseudo-principles (as a word can have different roles like noun, verb etc and also different contexts or meaning) and this creates a large spectrum of operation and valid possibilities, which can be narrowed only through understanding the interactions. While, we can only deal with the objects and the role transformation as our aiTPR, we provided evidence that the interaction terms are far more important and can help bridging the gap created by the object subspace, like in \cite{ren2015faster}, where the object space is mere structural from images or the features space is randomly defined to fit. 

\begin{figure*}[!h] 
\centering 
\includegraphics[width=\textwidth]{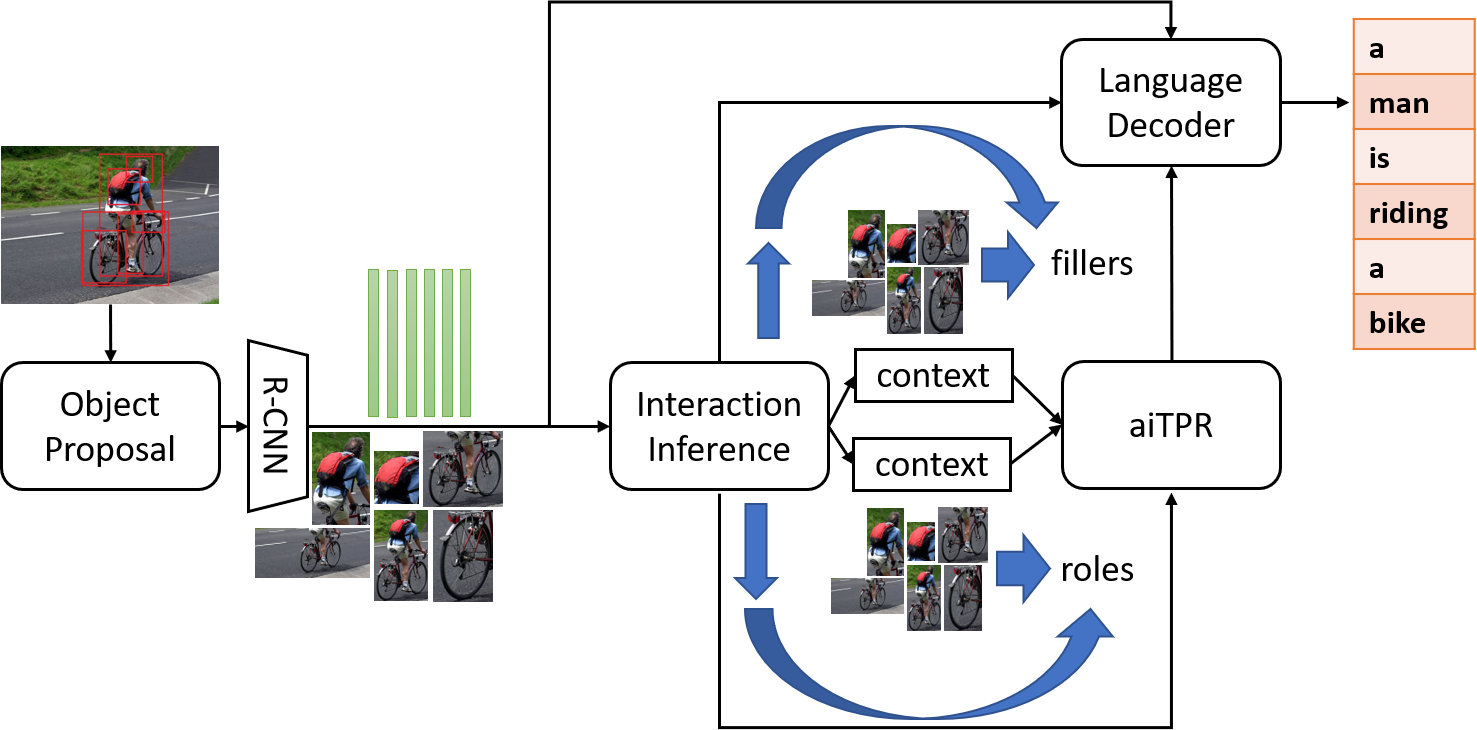}  
\caption{Overview Architecture of Attribute Interaction-Tensor Product Representation (aiTPR) as a Combination of Tensors Derived from Image Level Attributes and Inferred Interactions.}
\label{fig:Overview_architecture}
\end{figure*}

Previous works in image captioning focused on visual features from pre-trained network capable to detect objects with high precision like Vgg \cite{Karpathy2015Deep}, ResNet (\cite{ren2015faster, Devlin2015Language}), Inception \cite{vinyals2015show} etc, later gradually going multimodal (\cite{Karpathy2015Deep, Chen2015Mind, devlin2015exploring}) through different stages of feature, expecting that different layers will capture different aspects. Later on, gradually attention \cite{Xu2015Show, vinyals2015show, Mao2014deep, Devlin2015Language, yao2017boosting, rennie2017self, chen2018show}) based features became more popular due to the limitations of LSTM to keep its memory intact and limitations of the weighted transformations to capture all the relevant features required for diversified and correct attributed captions. Tensor Product Representation (TPR) concept is widely studied mathematical model with several mathematical properties, popular among people who wants to deal with orthogonality. However, the computational TPR space is yet to be explored. Recently, TPR \cite{smolensky1990tensor} is used to solve the question answering problem and the image captioning problem, where three models were proposed: 
dual generation model \cite{sur2019representations} for capturing both the context and roles for next word prediction, 
attention based model \cite{sur2019representations} for capturing the interdependence of the different words and by-passing the role of semantic features, and lastly the graphical tensor product representation model \cite{sur2019representations} combing both semantics and structural perceptions of triplet graphs for local features extraction and establishing the spatial temporal relationship for sentence generation. 
In this work, we will mostly concentrate on some novel structures for caption generation where the main focus is on categorically understand the individual components and produce most favorable representation for best reverse generation. The reverse generation is mostly done through language and some cases through images, like finding the most relevant images or generate images through GAN architectures. Reverse generation training has the problem of gathering biasness for the models and restricts generation of diverse sentences and out of box thinking or combination possibilities. Image captioning has always provided scope of further possibilities for video narration and understanding of sequentially related image spaces and story-telling. 

The rest of the document is arranged as Section \ref{section:literature} with the revisit of the existing works, descriptions of the problem in Section \ref{section:description}, methodology and architectural details in Section \ref{section:methodology}, experimental details, numerical and qualitative results and analysis in Section \ref{section:experiments} and conclusion and discussion of future works in  Section \ref{section:discussion}.

The main contribution of this paper are the followings: 1) defining a new architecture and concept of representation generation and fusion known as Attribute Interaction-Tensor Product Representation (aiTPR) 2) interactions bind different features together along with the orthogonality of the computational TPR architecture  3) our model contributed to the fact that orthogonality decides feature spaces and restricts mixing and mapping to undefined feature spaces 4) our model out-performed many previous works done on this domain and on the MSCOCO data.

\section{Description} \label{section:description}
The existing problem in language generation problem is the disability of the machines to differentiate between similar quite of situations, which arises because of the similar kind of combination of the aspects. In this work, our proposal architecture handles this problem in a different way, where we derive different levels of fine details of an image as likelihood, then the interactions are derived and then we derive different context and role generations for the sentence to be completed. However, compared to many previous works like in \cite{sur2019representations}, this approach is much more exploratory and performed better in caption generation. 

\subsection{Problem Description}
Generative application is gradually gained potential as we need to make the machine talk to humans. However, biasness in sentence generation held the topological exploration for a longer time. In this work, we explored the possibility of creating interaction criteria as an estimation ad proceed them with the regional image features for derivation of the representations for context and roles as we can call them as tensor product when we consider them as a combined representation ($\textbf{T}_{t}$ in Equation \ref{eq:tpr}). We deliberately create this tensor product so that we can multiply them as tensors and can define the semantics for the languages. 

\subsection{Difficulties Faced So Far}
Normal network learning does through these five difficulties apart from vanishing and exploding gradient problems, which is either bypassed or overlooked. 
These are as follows and we described here, how we can overcome this, mainly focusing on the model we have define in this work. 
\paragraph{Object and Combination Problem} 
When similar kinds of object combination exists in images, it becomes very difficult for the CNN network to differentiate them and provide an unique solution. In that regard, aiTPR can provide better solutions.
\paragraph{Weight Learning Problem}
While the weights are required to learn large part of the transformation for large amount of data, aiTPR can be a better alternative as it segregates the information in the form of a graphical structure instead of transformation.
\paragraph{Lower and Higher Level Understanding Problem} While the lower level features are the image features, several combinations created by aiTPR can help in identification of the unique opportunities for new attributes to crop up in sentences along with attributes. 
\paragraph{Summation to Undefined Spaces Problem} Summation problem is another problem that should be dealt with, mainly with the presence of large number of objects and their forms. Several such summation may converge them to similar spaces or to null spaces. 
\paragraph{Representation Approximation} Due to representation approximation, large part of the information are suppressed or gets mixed into unidentified states. To counter that, aiTPR keeps up the states through orthogonal transformation and are expected to regain the subspace, when required. 
All these when combined can provide a much better alternative and structurally sound neural network, that can be more explained.


\section{Our Methodology} \label{section:methodology}
Many organizations used an ensemble approach to avoid these problems instead of solving it naturally. In this work, we have proposed series of solutions for these above problems through use of other learned networks and intermediate inferences, which provides ample scope to neutralize the discrepancies in comprehension of the representations. Here, we have used extensively the region based object features and extracted different layer of information and fused them for representation generation in our network architecture and for each iteration, these representations took a new form so that the captions did not have to go through the biasness of non-linear approximations. 

\begin{figure*}[!h] 
\centering 
\includegraphics[width=\textwidth]{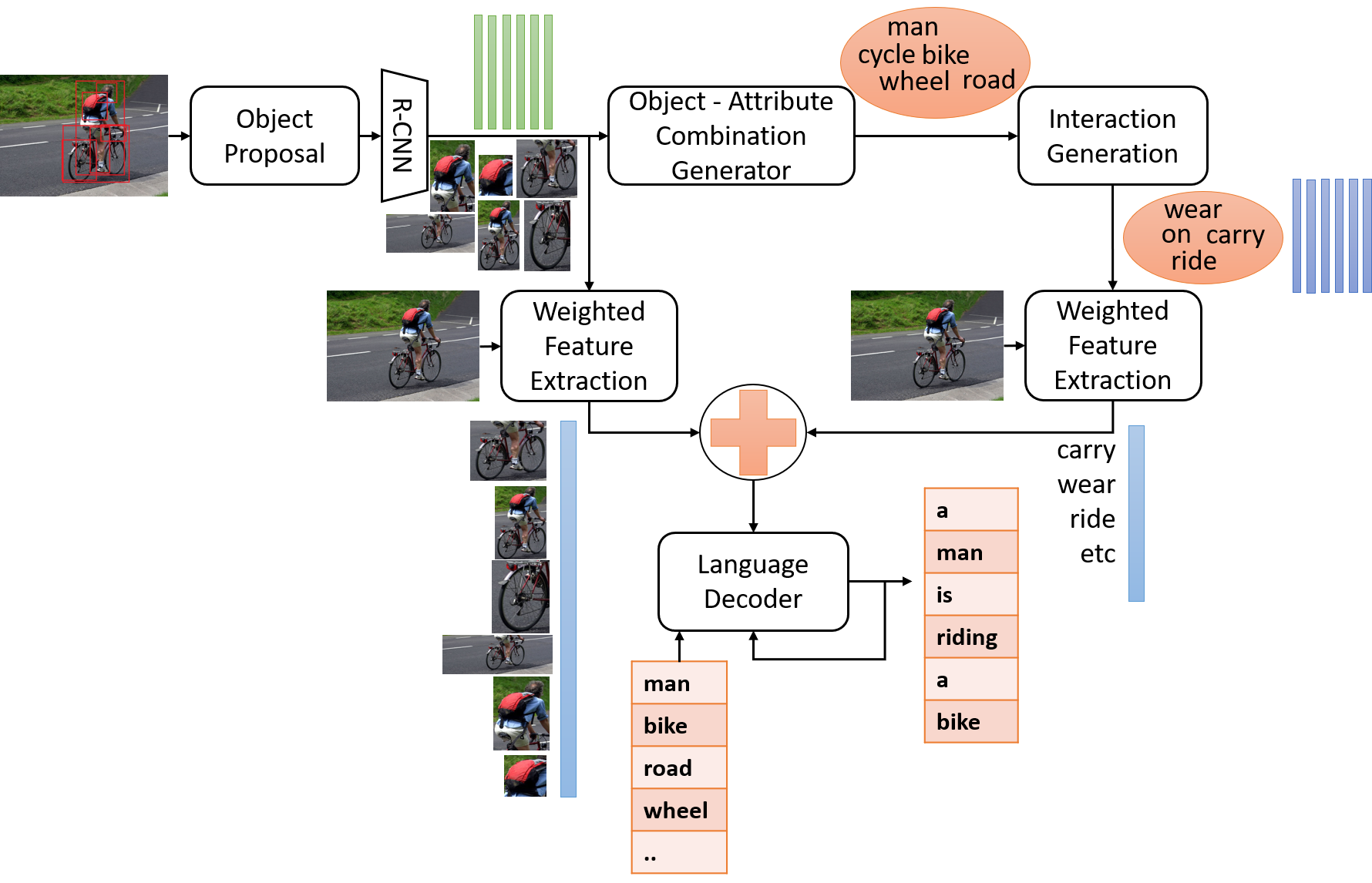}  
\caption{Early Attribute Interaction-Tensor Product Representation (aiTPR).}
\label{fig:Overall_architecture_late}
\end{figure*}

\begin{figure*}[!h] 
\centering 
\includegraphics[width=\textwidth]{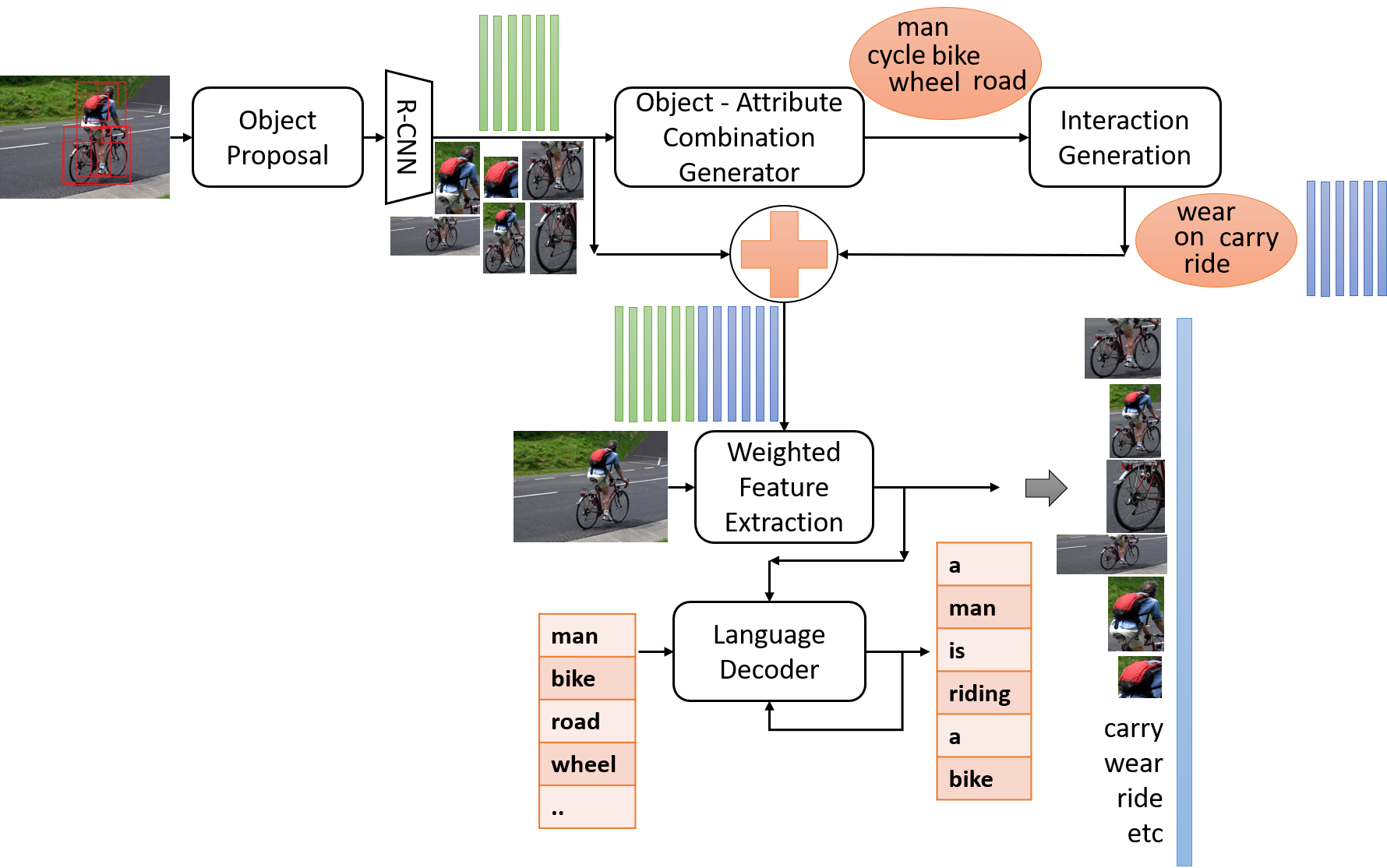}  
\caption{Late Attribute Interaction-Tensor Product Representation (aiTPR).}
\label{fig:Overall_architecture}
\end{figure*}

Here, we have provided the concept of Attribute Interaction-Tensor Product Representation (aiTPR) which operates on the regional features and their interaction criteria. While, the interactions are transformed replications of the objects features and are generated through inference, we can still consider two separate strategies to engage the attributes and the interaction segments. Figure \ref{fig:Overall_architecture_late} provided Late Attribute Interaction-Tensor Product Representation (aiTPR) while Figure \ref{fig:Overall_architecture} provided Early Attribute Interaction-Tensor Product Representation (aiTPR) architectures. The main difference is the way thenmixture of the objects representations takes place as a weighted sum of the lower level features of regional images. This kind of strategies are already in the literature like in MIMO antenna, where the aim is to estimate the best possible entropy from series of similar signals.
The novelty of our procedure is that we used the whole image and series of spatial relationships as a semantic composition denoting objects and activity-relationships, which provides opportunites for new objects and their interactions get more attention than before.

\subsection{Early Attribute Interaction-Tensor Product Representation} 
Early Attribute Interaction-Tensor Product Representation uses the strategy of linking the attribute and interaction tensors much early in the network, considering the fact that there will be correlations establishment among them much earlier. 
Mathematically, we represent Early Attribute Interaction-Tensor Product Representation with the following set of equations. If we define $v \in \mathbb{R}^{2048}$ as the visual feature for image $\textbf{I}$ and $\textbf{v} = \{v_1,\ldots,v_n\}$ as the RCNN based region based object feature representation where each $v_i \in \mathbb{R}^{2048}$ for $i \in \{1,\ldots,n\}$ and $n$ is the number of regional objects detected in the image $\textbf{I}$ making the overall $\textbf{v} \in \mathbb{R}^{n \times 2048}$. 
The initial parameters for the Assembled Selector Layer are initialized as the followings, considering that the biasness of the network must be neutralized and it will also help in estimation of the content of the images. 
\begin{equation}
 \overline{v} = \frac{1}{k} \sum\limits_{i=1}^{i=k} v_i
\end{equation}
\begin{equation}
 \overline{v} = \textbf{v}
\end{equation}
The initial parameters are initialized as the followings. 
\begin{equation}
 \textbf{h}_{0}, \textbf{ } \textbf{c}_{0} = \textbf{W}_{h_0}\overline{v}, \textbf{W}_{c_0}\overline{v}
\end{equation}
$\textbf{W}_{h_0} \in \mathbb{R}^{2048 \times d}$, $\textbf{W}_{c_0} \in \mathbb{R}^{2048 \times d}$.  
The Intermediate Transfer Layer helps in transferring the 
\begin{equation} \label{eq:st1}
 \textbf{a}_{t} = \textbf{W}_{a} \tanh (\textbf{W}_{h} \textbf{h}_{t-1})
\end{equation}
where $\textbf{W}_{a} \in \mathbb{R}^{b \times d}$, $\textbf{W}_{h} \in \mathbb{R}^{d \times k}$.
\begin{equation}
 \alpha_t = \mathrm{softmax}(\textbf{a}_t)
\end{equation}
$\textbf{a}_t \in \mathbb{R}^{k} \in \{a_{1,t},\ldots,a_{k,t}\}$
\begin{equation} \label{eq:en1}
 \widehat{v}_t =  \frac{\left[ \sum\limits_{i=1}^{i=k_1} v_i \alpha_{i,t} \text{ }\text{ }  +  \text{ }\text{ } \sum\limits_{i=1}^{i=k_2} v'_i \alpha'_{i,t} \right]}{2}
\end{equation}
with $k = k_1 + k_2$, $\sum \alpha_{i} = 1$ and $\sum \alpha_{i} = 1$ $v$ is the regional CNN and $v'$ is the representation of the objects detected through the regional CNN model. $\widehat{v}_t \in \mathbb{R}^{b \times d}$ where $b$ is the batch size and $d$ is the hidden layer dimension. 
The Assembled Selector Layer ca be denoted as the following equations,
\begin{equation}
 \textbf{q}_{t} = \widehat{v}_t
\end{equation}
\begin{equation}
 \textbf{p}_{t} = \textbf{W}_e \textbf{x}_{t-1}
\end{equation}
\begin{equation}
\begin{split}
 \textbf{T}_{t} & =  
 \textbf{W}_{s_{12}}\text{ }\sigma(\textbf{W}_{s_{11}}\textbf{h}_{t-1} + \textbf{b}_1) \\
 & \otimes 
 \tanh (\textbf{W}_{s_{22}} ( \textbf{v}_x\text{ }\sigma(\textbf{W}_{s_{21}}\textbf{h}_{t-1} + \textbf{b}_2) ) + \textbf{b}_3) 
\end{split}
\end{equation}
\begin{equation}
\begin{split}
 \textbf{T}_{t} & =  
 \textbf{W}_{s_{12}}\text{ }\sigma(\textbf{W}_{s_{11}}\textbf{h}_{t-1} + \textbf{W}_{w_{_{1}}}\sum\limits_{i=0}^{t-1} \textbf{W}_e \textbf{x}_{i} + \textbf{b}_1) 
 \text{ }\otimes \\
 & \tanh (\textbf{W}_{s_{22}} ( \textbf{v}_x\text{ }\sigma(\textbf{W}_{s_{21}}\textbf{h}_{t-1} + \textbf{W}_{w_{_{2}}}\sum\limits_{i=0}^{t-1} \textbf{W}_e \textbf{x}_{i} + \textbf{b}_2) ) + \textbf{b}_3) 
\end{split}
\end{equation}
\begin{equation}
 \textbf{v}_x = f_x(\{v_{a_1},v_{a_2},\ldots,v_{b_1},v_{b_2},\ldots \})
\end{equation}
where we have $f_x(.)$ as a function and  $v_{a_i} \in \mathbb{R}^{2048} $ and $v_{b_i} \in \mathbb{R}^{2048}$ are the attribute and interaction components. 
$\otimes$ is an algebraic operation. Here, we considered $\otimes = \odot$ as we try to rectify one context with the other context. 
\begin{equation} \label{eq:tag1a}
\textbf{q}_{t} = \textbf{W}_{h,m} S \odot \textbf{W}_{h,n} \textbf{q}_{t}
\end{equation}
\begin{equation} \label{eq:tag1b}
\textbf{p}_{t} = \textbf{W}_{h,m} S \odot \textbf{W}_{h,n} \textbf{p}_{t}
\end{equation}
\begin{equation}
 \textbf{i}_{t} = \sigma(\textbf{W}_{pi}\textbf{p}_{t} + \textbf{W}_{qi}\textbf{q}_{t} + \textbf{W}_{Ti}\textbf{T}_{t} + \textbf{b}_{i})
\end{equation}
\begin{equation}
 \textbf{f}_{t} = \sigma(\textbf{W}_{pf}\textbf{p}_{t} + \textbf{W}_{qf}\textbf{q}_{t} + \textbf{W}_{Tf}\textbf{T}_{t} + \textbf{b}_{f})
\end{equation}
\begin{equation}
 \textbf{o}_{t} = \sigma(\textbf{W}_{po}\textbf{p}_{t} + \textbf{W}_{qo}\textbf{q}_{t} + \textbf{W}_{To}\textbf{T}_{t} + \textbf{b}_{o})
\end{equation}
\begin{equation}
 \textbf{g}_{t} = \tanh(\textbf{W}_{pg}\textbf{p}_{t} + \textbf{W}_{qg}\textbf{q}_{t} + \textbf{W}_{Tg}\textbf{T}_{t} + \textbf{b}_{g})
\end{equation}
\begin{equation}
 \textbf{c}_{t} = \textbf{f}_{t} \odot \textbf{c}_{t-1} + \textbf{i}_{t} \odot \textbf{g}_{t} 
\end{equation}
\begin{equation}
 \textbf{h}_{t} = \textbf{o}_{t} \odot \tanh(\textbf{c}_{t})
\end{equation}
\begin{equation}
 \textbf{x}_{t} = \max \arg \mathrm{softmax} (\textbf{W}_{hx} \textbf{h}_{t})
\end{equation}

Mathematically, Early Attribute Interaction-Tensor Product Representation (aiTPR), denoted as $f_{_{aiTPR_E}}(.)$, can be described as the followings probability distribution estimation.
\begin{equation} \label{eq:itbe}
\begin{split}
 f & _{_{aiTPR_E}}(\textbf{v}) = \prod\limits_{k}^{} \mathrm{Pr}(w_k \mid \textbf{T}_i, \text{ } \textbf{v},\text{ } \textbf{W}_{L_1}) \\
 &  \prod\limits_{i}^{} \mathrm{Pr}(\textbf{T}_i \mid \textbf{v},  \text{ }\textbf{W}_1)  \\
 & = \prod\limits_{k}^{} \mathrm{Pr}(w_k \mid \textbf{T}_i, \text{ } \left( \frac{1}{K}\sum\limits_{m=1}^{K} v_m \right),  \frac{\left( \sum\limits_{m=1}^{N_1} {a}_mv_m \right)}{2} \text{ } + \text{ }  \\
 & \text{ }\text{ } \frac{\left( \sum\limits_{m=1}^{N_2} {a'}_mv'_m \right)}{2}, \text{ } \textbf{W}_{L_1} ) \prod\limits_{i}^{} \mathrm{Pr}(\textbf{T}_i \mid \textbf{v},  \text{ }\textbf{W}_1)  \\
 & = \prod\limits_{k}^{} \mathrm{Q}_{IC}(w_k \mid \textbf{T}_i, \text{ } \left( \frac{1}{K}\sum\limits_{m=1}^{K} v_m \right),   \frac{ \left( \sum\limits_{m=1}^{N_1} {a}_mv_m \right)}{2} \text{ } + \text{ }  \\
 & \text{ }\text{ } \frac{\left( \sum\limits_{m=1}^{N_2} {a'}_mv'_m \right)}{2} ) \prod\limits_{i}^{} \mathrm{Q}(\textbf{T}_i \mid \textbf{v}) 
\end{split} 
\end{equation}
using the weights of the LSTM in the architecture is denoted as $\textbf{W}_{L_1}$, $w_i$ as words of sentences, $v_i$ as regional image features, ${a}_ms_m$ as intermediate learnt parameters, $\mathrm{Q}_{IC}(.)$ and $\mathrm{Q}(.)$ are the Image Caption and Scene-Graph generator function respectively. $\mathrm{Q}(.)$ derives $\textbf{x}$ (Scene-Graph information) from $\textbf{v}$ of $\textbf{I}$.
These set of equations worked best for this kinds of situation and experimented it for understanding the effects of our new concept of Attribute Interaction-Tensor Product Representation (aiTPR). Apart from that, this network architecture helps in fusion of different set of feature tensors without the scope of  influence or suppression and compared to architectures like \cite{lu2018neural}, it is much lighter in the number of weights.

\subsection{Late Attribute Interaction-Tensor Product Representation} 
Late Attribute Interaction-Tensor Product Representation (aiTPR) kept the attribute and interaction separated so that the both the network learns equivalently before they fused for the language decoder. The Late aiTPR model operates without any expectation of correlation between the attribute set and the interaction set. While, most of the works in attention are much concentrated on automatic segregation of useful information, we propose that the features must segregate in pure form before getting into the series of linear and non-linear approximations. 
The contexts from regional object features are coupled along, just like tensor product, but instead of word embedding, some of the attributes are still the image features from a RCNN network. Before, we discuss the Late aiTPR, we revisit the concept of TPR as a general. 

\subsubsection{TPR}
For a TPR, we have Here, the context is represented as $f_i \in \mathbb{R}^{d_n}$ and the topological information vector as $r_i \in \mathbb{R}^{d_m}$ creating the summation of orthogonal vectors as $\sum\limits_{}^{} f_ir_i$ and using the same topological information vector, 
we get back $f_j$ as $f_j = r_j^T \sum\limits_{}^{} f_ir_i  = f_1r_1r_j^T + f_2r_2r_j^T + \ldots = f_jr_jr_j^T = f_j$. here $f$ is object features and $r$ is orthogonal vectors where both represent an output in caption. The novelty lies in a direct relationship between object to word generation without explicitly knowing the identity of the words. It is aided by the other learnable weights. TPR can be represented as $\textbf{s}(\textbf{w})$ as, 
\begin{equation}
 \textbf{s}(\textbf{w}) = \sum f_i \otimes r^T_i
\end{equation}
where $\textbf{w}$ is the feature vector, and $\{\textbf{w} \rightarrow \textbf{f} : \textbf{w} \in \textbf{W}_{e} \}$ is the transformation, $\textbf{W}_{e}$ are the raw features or the embedding vectors for features which minimizes the context function

Mathematically, the equations that will describe the architecture can be introduced as the followings, 
\begin{equation}
 \overline{v} = \frac{1}{k} \sum\limits_{i=1}^{i=k} v_i
\end{equation}
\begin{equation}
 \overline{v} = \textbf{v}
\end{equation}
The initial parameters are initialized as the followings. 
\begin{equation}
 \textbf{h}_{0}, \textbf{ } \textbf{c}_{0} = \textbf{W}_{h_0}\overline{v}, \textbf{W}_{c_0}\overline{v}
\end{equation}
where we have $ \textbf{W}_{h_0} \in \mathbb{R}^{2048 \times d}$, $ \textbf{W}_{c_0} \in \mathbb{R}^{2048 \times d}$.
The next segment equations can be denoted as the following,  
\begin{equation} \label{eq:st1}
 \textbf{a}_{t} = \textbf{W}_{a} \tanh (\textbf{W}_{h} \textbf{h}_{t-1})
\end{equation}
where $\textbf{W}_{a} \in \mathbb{R}^{b \times d}$, $\textbf{W}_{h} \in \mathbb{R}^{d \times k}$.
\begin{equation}
 \alpha_t = \mathrm{softmax}(\textbf{a}_t)
\end{equation}
$\textbf{a}_t \in \mathbb{R}^{k} \in \{a_{1,t},\ldots,a_{k,t}\}$
\begin{equation} \label{eq:en1}
 \widehat{v}_t =  \left[ \sum\limits_{i=1}^{i=k_1} v_i \alpha_{i,t} \text{ }\text{ }  +  \text{ }\text{ } \sum\limits_{i=1}^{i=k_2} v'_i \alpha_{i,t} \right]
\end{equation}
$\sum \alpha_{i}  = 1$, 
$\widehat{v}_t \in \mathbb{R}^{b \times d}$ where $b$ is the batch size and $d$ is the hidden layer dimension. 
\begin{equation}
 \textbf{q}_{t} = \widehat{v}_t
\end{equation}
\begin{equation}
 \textbf{p}_{t} = \textbf{W}_e \textbf{x}_{t-1}
\end{equation}
\begin{equation}
\begin{split}
 \textbf{T}_{t} & =  
 \textbf{W}_{s_{12}}\text{ }\sigma(\textbf{W}_{s_{11}}\textbf{h}_{t-1} + \textbf{b}_1) \\
 & \otimes 
 \tanh (\textbf{W}_{s_{22}} ( \textbf{v}\text{ }\sigma(\textbf{W}_{s_{21}}\textbf{h}_{t-1} + \textbf{b}_2) ) + \textbf{b}_3) 
\end{split}
\end{equation}
\begin{equation} \label{eq:tpr}
\begin{split}
 \textbf{T}_{t} & =  
 \textbf{W}_{s_{12}}\text{ }\sigma(\textbf{W}_{s_{11}}\textbf{h}_{t-1} + \textbf{W}_{w_{_{1}}}\sum\limits_{i=0}^{t-1} \textbf{W}_e \textbf{x}_{i} + \textbf{b}_1) 
 \text{ }\otimes \\
 & \tanh (\textbf{W}_{s_{22}} ( \textbf{v}\text{ }\sigma(\textbf{W}_{s_{21}}\textbf{h}_{t-1} + \textbf{W}_{w_{_{2}}}\sum\limits_{i=0}^{t-1} \textbf{W}_e \textbf{x}_{i} + \textbf{b}_2) ) + \textbf{b}_3) 
\end{split}
\end{equation}
$\otimes$ is an algebraic operation. Here, we considered $\otimes = \odot$ as we try to rectify one context with the other context. 
\begin{equation} \label{eq:tag2a}
\textbf{q}_{t} = \textbf{W}_{h,m} S \odot \textbf{W}_{h,n} \textbf{q}_{t}
\end{equation}
\begin{equation} \label{eq:tag2b}
\textbf{p}_{t} = \textbf{W}_{h,m} S \odot \textbf{W}_{h,n} \textbf{p}_{t}
\end{equation}
\begin{equation}
 \textbf{i}_{t} = \sigma(\textbf{W}_{pi}\textbf{p}_{t} + \textbf{W}_{qi}\textbf{q}_{t} + \textbf{W}_{Ti}\textbf{T}_{t} + \textbf{b}_{i})
\end{equation}
\begin{equation}
 \textbf{f}_{t} = \sigma(\textbf{W}_{pf}\textbf{p}_{t} + \textbf{W}_{qf}\textbf{q}_{t} + \textbf{W}_{Tf}\textbf{T}_{t} + \textbf{b}_{f})
\end{equation}
\begin{equation}
 \textbf{o}_{t} = \sigma(\textbf{W}_{po}\textbf{p}_{t} + \textbf{W}_{qo}\textbf{q}_{t} + \textbf{W}_{To}\textbf{T}_{t} + \textbf{b}_{o})
\end{equation}
\begin{equation}
 \textbf{g}_{t} = \tanh(\textbf{W}_{pg}\textbf{p}_{t} + \textbf{W}_{qg}\textbf{q}_{t} + \textbf{W}_{Tg}\textbf{T}_{t} + \textbf{b}_{g})
\end{equation}
\begin{equation}
 \textbf{c}_{t} = \textbf{f}_{t} \odot \textbf{c}_{t-1} + \textbf{i}_{t} \odot \textbf{g}_{t} 
\end{equation}
\begin{equation}
 \textbf{h}_{t} = \textbf{o}_{t} \odot \tanh(\textbf{c}_{t})
\end{equation}
\begin{equation}
 \textbf{x}_{t} = \max \arg \mathrm{softmax} (\textbf{W}_{hx} \textbf{h}_{t})
\end{equation}

Mathematically, Late Attribute Interaction-Tensor Product Representation (aiTPR), denoted as $f_{_{aiTPR_L}}(.)$, can be described as the followings probability distribution estimation.
\begin{equation} \label{eq:itbl}
\begin{split}
 f & _{_{aiTPR_L}}(\textbf{v}) = \prod\limits_{k}^{} \mathrm{Pr}(w_k \mid \textbf{T}_i, \text{ } \textbf{v},\text{ } \textbf{W}_{L_1}) \\
 &  \prod\limits_{i}^{} \mathrm{Pr}(\textbf{T}_i \mid \textbf{v},  \text{ }\textbf{W}_1)  \\
 & = \prod\limits_{k}^{} \mathrm{Pr}(w_k \mid \textbf{T}_i, \text{ } \left( \frac{1}{K}\sum\limits_{m=1}^{K} v_m \right), \left( \sum\limits_{m=1}^{N_1} {a}_mv_m \right) \text{ } + \text{ }  \\
 & \text{ }\text{ } \left( \sum\limits_{m=1}^{N_2} {a'}_mv'_m \right), \text{ } \textbf{W}_{L_1} ) \prod\limits_{i}^{} \mathrm{Pr}(\textbf{T}_i \mid \textbf{v},  \text{ }\textbf{W}_1)  \\
 & = \prod\limits_{k}^{} \mathrm{Q}_{IC}(w_k \mid \textbf{T}_i, \text{ } \left( \frac{1}{K}\sum\limits_{m=1}^{K} v_m \right),  \left( \sum\limits_{m=1}^{N_1} {a}_mv_m \right)\text{ } + \text{ }  \\
 & \text{ }\text{ }\left( \sum\limits_{m=1}^{N_2} {a'}_mv'_m \right) ) \prod\limits_{i}^{} \mathrm{Q}(\textbf{T}_i \mid \textbf{v}) 
\end{split} 
\end{equation}
using the weights of the LSTM in the architecture is denoted as $\textbf{W}_{L_1}$, $w_i$ as words of sentences, $v_i$ as regional image features, ${a}_ms_m$ as intermediate learnt parameters, $\mathrm{Q}_{IC}(.)$ and $\mathrm{Q}(.)$ are the Image Caption and TPR generator function respectively.

\subsection{Uniqueness of aiTPR}
The most components and uniqueness of our architecture is the introduction and fusion of regional image features and several other abstracted likelihood embedding of interaction terms. Unlike other works, where they have used image feature, a different kind of constructed features can bring stability in the variance of the features and can restrict the null spaces. While, many works have demonstrated regional image features based networks, they have left out the scope of introducing the fusion points of these regional influences. In this work, we have introduced such a scheme called interactions and have presented in the form of TPR, where we construct two different contexts from the same feature space and use them as product for maximum influence.

\section{Experiments, Results \& Analysis} \label{section:experiments}
We have done wide range of experiments to show the behavioral influence of object based attention through this architectural network, where we have defined different levels of information for the generation of captions. Before we analyze the results, short description of the dataset is also provided along with the training session description as we explore the joint distribution of the learning state space.  

\subsection{Data Description}
We have used the MSCOCO and the Visual Genome dataset for our analysis. MSCOCO consists of 123287 train+validation images and 566747 train+validation sentence, where each image is associated with at least five sentences from a vocabulary of 8791 words. There are 5000 images (with 25010 sentences) for validation and 5000 images (with 25010 sentences) for testing. We used the same data split as described in Karpathy's paper \cite{Karpathy2015Deep}. Visual Genome dataset is used for other language semantic information for the MSCOCO datasets and a model is trained to derive the annotations for those images. Roughly, 38\% of the MSCOCO data has attribute level annotations in the Visual Genome dataset.

\subsection{Quantitative Evaluation}
Several evaluation metrics like CIDEr-D, Bleu\_4,  Bleu\_3, Bleu\_2, Bleu\_1, ROUGE\_L, METEOR and SPICE is used for our experiments. Table \ref{table:performanceEva} provide a quantitative evaluation of our experiments, mainly focusing on the different architectures, related to the context $\textbf{h}_{t-1}$ and previous word embedding $\textbf{x}_{t-1}$ semantics, which is often found to enhance the performance for the captions. 
We found that the performance of our model out-performed many previous models and it was also found that the late fusion model provide better performance, inducing more experiments based on the semantic correction. Clearly, Late aiTPR  [aiTPR (3)] with Equation \ref{eq:tag2a} was the winner interms of most of the evaluation metrics, but the other schemres like [aiTPR (1)] and [aiTPR (2)] performed in a competitive way. The main reason, [aiTPR (3)] worked better was because of the influence it produced on the combinations of regional feature set and their interactions, which ultimately helped in better captions, while Equation \ref{eq:tag2b} helped in the topological continuity of the word for the generated sentence. 

\begin{table*}
\centering
\caption{Performance Evaluation for Different Architectures without Reinforcement Learning}
\begin{tabular}{|c|c|c|c|c|c|c|c|c|}
\hline
 Algorithm & CIDEr-D & Bleu\_4 & Bleu\_3 & Bleu\_2 & Bleu\_1 & ROUGE\_L & METEOR  &  SPICE \\ 
\hline \hline
 Human \cite{wu2017image} & 0.85 & 0.22 & 0.32 & 0.47 & 0.66 & 0.48 & 0.2 & -- \\ 
    Neural Talk \cite{Karpathy2015Deep} & 0.66 & 0.23 & 0.32 & 0.45 & 0.63 & -- & 0.20 & --  \\ 
    Mind’sEye \cite{Chen2015Mind} & -- & 0.19 & -- & -- & -- & -- & 0.20 & --  \\
    Google \cite{vinyals2015show}  & 0.94 & 0.31 & 0.41 & 0.54 & 0.71 & 0.53 & 0.25 & --  \\
    LRCN \cite{Donahue2015Long-term} & 0.87 & 0.28 & 0.38 & 0.53 & 0.70 & 0.52 & 0.24 & --  \\   
    Montreal \cite{Xu2015Show} & 0.87 & 0.28 & 0.38 & 0.53 & 0.71 & 0.52 &  0.24 & -- \\ 
    m-RNN \cite{Mao2014deep}  & 0.79 & 0.27 & 0.37 & 0.51 & 0.68 & 0.50 & 0.23 & --  \\
    \cite{Jia2015}  & 0.81 & 0.26 & 0.36 & 0.49 & 0.67 & -- & 0.23 & --  \\
    MSR \cite{Fang2015captions} & 0.91 & 0.29 & 0.39 & 0.53 & 0.70 & 0.52 & 0.25 & --  \\
    \cite{Jin2015Aligning} & 0.84 & 0.28 & 0.38 & 0.52 & 0.70 & -- & 0.24 & --  \\
    bi-LSTM \cite{wang2018image} & -- & 0.244 & 0.352 & 0.492 & 0.672 & -- & -- & --  \\ 
    MSR Captivator \cite{Devlin2015Language} & 0.93 & 0.31 & 0.41 & 0.54 & 0.72 & 0.53 & 0.25 & -- \\
    Nearest Neighbor \cite{devlin2015exploring} & 0.89 & 0.28 & 0.38 & 0.52 & 0.70 & 0.51 & 0.24 & -- \\
    ATT \cite{You2016Image} & 0.94 & 0.32 & 0.42 & 0.57 & 0.73 & 0.54 & 0.25 & -- \\
    \cite{wu2017image} & 0.92 & 0.31 & 0.41 & 0.56 & 0.73 & 0.53 & 0.25 & -- \\
    \hline\hline
    Adaptive \cite{lu2017knowing}  & 1.085 & 0.332 & 0.439 & 0.580 & 0.742 & -- & 0.266 & --  \\
    MSM \cite{yao2017boosting}  & 0.986 & 0.325 & 0.429 & 0.565 & 0.730 & -- & 0.251 & --  \\ 
    ERD \cite{yang1605encode}  & 0.895 & 0.298 & -- & -- & -- & -- & 0.240 & --  \\ 
    Att2in \cite{rennie2017self}  & 1.01 & 0.313 & -- & -- & -- & -- & 0.260 & --  \\  
    Top-Down$\dagger$ \cite{anderson2018bottom}  & 1.054 & 0.334 & -- & -- & 0.745 & -- & 0.261 & 0.192  \\ %
    
    Attribute-Attention \cite{chen2018show} & 1.044 & 0.338 & 0.443 & 0.579 & 0.743 & 0.549  & -- & -- \\ %
    LSTM \cite{Gan2016} & 0.889 & 0.292 & 0.390 & 0.525 & 0.698 & -- & 0.238 & -- \\
    SCN \cite{Gan2016} & 1.012 & 0.330 & 0.433 & 0.566 & 0.728 & -- & 0.257 & --  \\

 NBT$\dagger$ \cite{lu2018neural}  & 1.07 & 0.347 & -- & -- & 0.755 & -- & 0.271 & 0.201  \\ 
 Top-Down$\dagger$ \cite{anderson2018bottom}  & 1.135 & 0.362 & -- & -- & 0.772 & 0.564 & 0.27 & 0.203  \\ %
 
  \hline \hline 
 
 Early aiTPR  with Equation \ref{eq:tag2a} & 1.387 & 0.476 & 0.588 & 0.718 & 0.850 & 0.632 & 0.318 & 0.233  \\
 
 Late aiTPR  with Equation \ref{eq:tag2a} & 1.401 & 0.484 & 0.595 & 0.721 & 0.850 & 0.637 & 0.320 & 0.233 \\
 
 \hline \hline 
 
  Late aiTPR  [aiTPR (1)] & 1.386 & 0.476 & 0.586 & 0.714 & 0.846 & 0.631 & 0.318 & 0.232 \\
  without Equation \ref{eq:tag2a} and Equation \ref{eq:tag2b} &  &  &  &  &  &  &  &  \\

  Late aiTPR  [aiTPR (2)]  & 1.398 & 0.484 & 0.593 & 0.721 & 0.852 & 0.635 & 0.320 & 0.233 \\
  with Equation \ref{eq:tag2a} and Equation \ref{eq:tag2b} &  &  &  &  &  &  &  &  \\

  Late aiTPR  [aiTPR (3)] & \textbf{1.401} & \textbf{0.484} & \textbf{0.595} & \textbf{0.721} & \textbf{0.850} & \textbf{0.637} & \textbf{0.320} & \textbf{0.233} \\
  with Equation \ref{eq:tag2a} &  &  &  &  &  &  &  &  \\

 \hline
 \multicolumn{9}{l}{\textsuperscript{$\dagger$}\footnotesize{Ensemble \& Reinforcement Learning Used with RCNN features and \cite{anderson2018bottom} Used 10K Vocabulary Dataset}}
\end{tabular}
\label{table:performanceEva}
\end{table*}

\subsection{Discussion}
The main improvement that our models had put into the architecture is the introduction of aiTPR, which is characterized by understanding the interaction level information based representation. This was never tried before and introduction of regional influences attended refined feature levels and provided ample scope of a fitted structure with improvement of the representations with iterations. While, previous works were concentrated on defining better image feature quality, we have paid more importance on the inference level information that generalizes the representations, but create combination level enhancements. While, most of the works were just what  the model has learned, we paid more mportance what we can create and feed into the network like shown in Equation \ref{eq:itbe} and Equation \ref{eq:itbl}. With this approach, we have establish a new performance level, which has surpassed other previous works in all the metrics. We have used a RCNN network to find the regional level details along with the coordinates of the regions and used them for interaction level inference without much concerning about the correctness as we are more interested in the defined representation than the inference level likelihood tensors.

\subsection{Qualitative Evaluation}
\begin{figure*}[t] 
\centering 
\includegraphics[width=\textwidth]{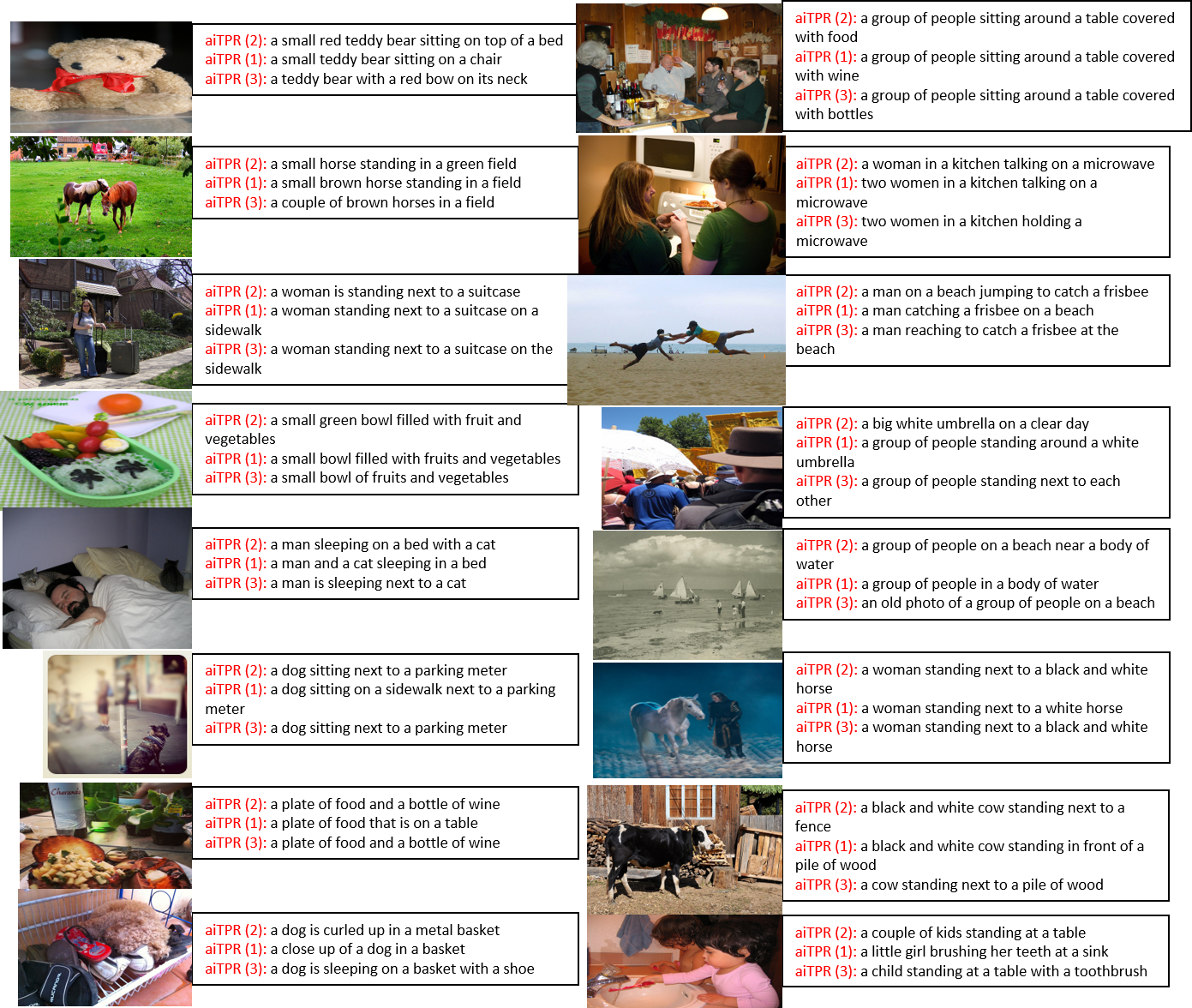}  
\caption{Qualitative Analysis. Part 1. [aiTPR (1)], [aiTPR (2)] and [aiTPR (3)] are defined Table \ref{table:performanceEva}.}
\label{fig:QualitativeAnalysis1}
\end{figure*}

\begin{figure*}[!h] 
\centering 
\includegraphics[width=\textwidth]{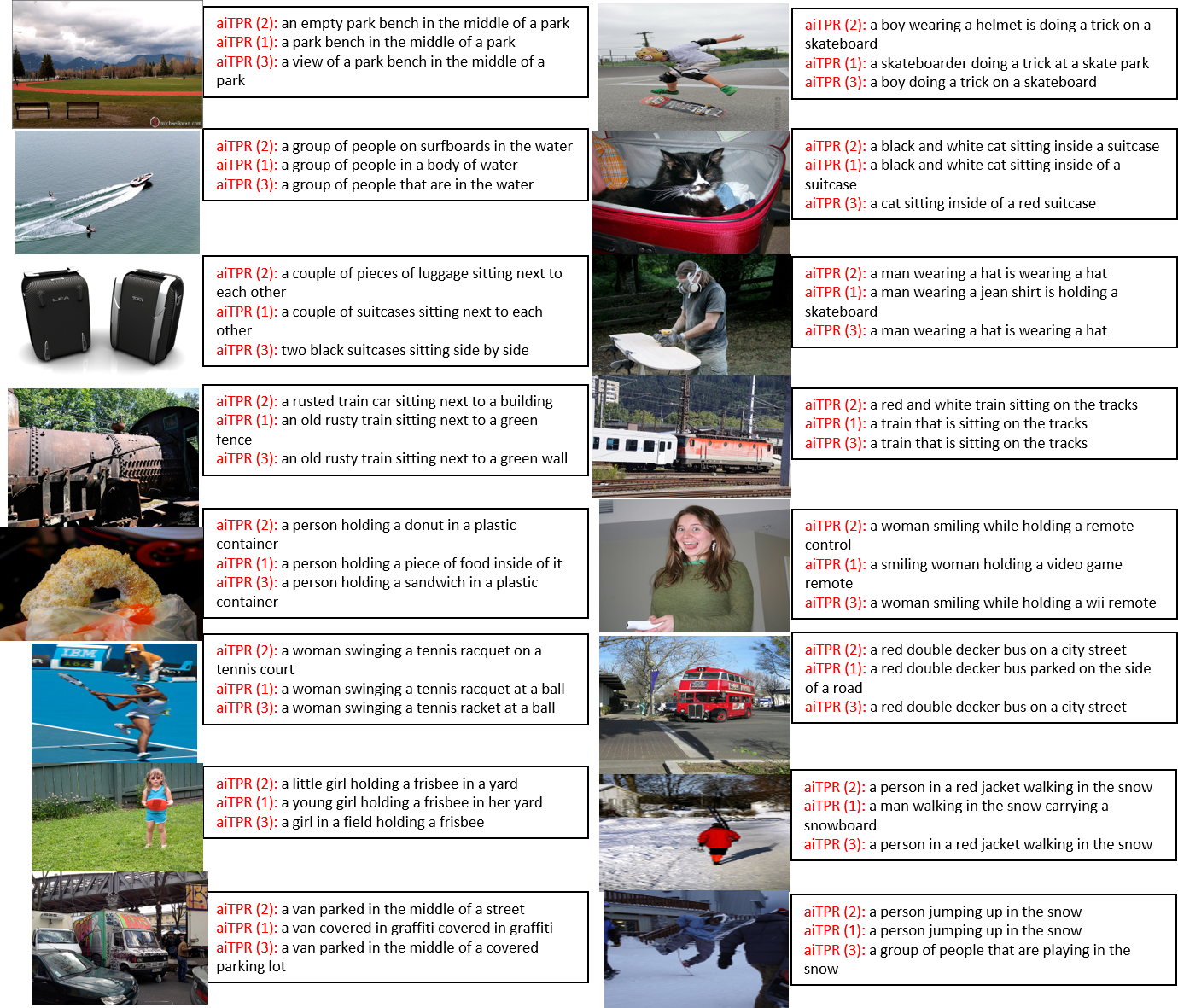}  
\caption{Qualitative Analysis. Part 2. [aiTPR (1)], [aiTPR (2)] and [aiTPR (3)] are defined Table \ref{table:performanceEva}.}
\label{fig:QualitativeAnalysis2}
\end{figure*}

Normally statistical formulas are the best evaluation of the significance levels of any model, except language ones. This is because the statistical models most concentrate on the content and the numericals associated with it.  To evaluate language structures, we need perception, which is also diverse and subject to high variations. Also, whether a model is better than the other cannot be judged through average numericals. Whether overall improved captions are generated or not is also difficult to judge from numericals in Table \ref{table:performanceEva}, Hence, we have considered some of the generated sentences as our qualitative analysis and will reflect the supremacy of our novel architecture. Figure \ref{fig:QualitativeAnalysis1} and Figure \ref{fig:QualitativeAnalysis2} provided some of the instances that were derived from the models. The examples provided very good representations of the generated sentences from the corresponding images. 


\section{Discussion} \label{section:discussion}
While the previous works mainly concentrated on the features and their combinations in an un-thoughtful way, this work produces a technique where you can derive useful interactions for the attributes and generate the most useful tensors and their products, without the requirement for non-linear approximation. We introduced, for the first time, the notion of combining regional image features and abstracted interaction likelihood embedding for image captioning. 
We call this model as Attribute Interaction-Tensor Product Representation (aiTPR) as this is an good AI technique to consider the attribute-interaction (ai) where the attributes are structured ones as the lower level features transferred from any pre-trained model, while the interactions are the composed derived from them and then approximated through a language representation that is well fitted with the image feature model. With this work, we have derived a new state-of-the-art result and also a novel work. 

\section*{Acknowledgment}
The author has used University of Florida HiperGator, equipped with NVIDIA Tesla K80 GPU,  extensively for the experiments. The author acknowledges University of Florida Research Computing for providing computational resources and support that have contributed to the research results reported in this publication. URL: http://researchcomputing.ufl.edu

\section*{Conflict of Interest} The author declares that he has no conflict of interest.

\ifCLASSOPTIONcaptionsoff
  \newpage
\fi

%








\end{document}